\documentclass{bmvc2k}

\title{ORA3D: Overlap Region Aware Multi-view 3D Object Detection}

\addauthor{Wonseok Roh}{paulroh@korea.ac.kr}{1}
\addauthor{Gyusam Chang}{gsjang95@korea.ac.kr}{1}
\addauthor{Seokha Moon}{shmoon96@korea.ac.kr}{3}
\addauthor{Giljoo Nam}{namgiljoo@gmail.com}{2}
\addauthor{Chanyoung Kim}{kochanha@gmail.com}{1}
\addauthor{Younghyun Kim}{yhkim84@hyundai.com}{4}
\addauthor{Jinkyu Kim}{jinkyukim@korea.ac.kr}{3*}
\addauthor{Sangpil Kim}{spk7@korea.ac.kr}{1*}

\addinstitution{
 Department of Artificial Intelligence,\\
 Korea University,\\
 Seoul, Republic of Korea
}
\addinstitution{
 School of Computing,\\
 KAIST,\\
 Daejeon, Republic of Korea
}
\addinstitution{
 Department of Computer Science and Engineering,\\
 Korea University,\\
 Seoul, Republic of Korea
}
\addinstitution{
 Autonomous Driving Center,\\
 Hyundai Motor Company R\&D Division,\\
 Seoul, Republic of Korea
}

\runninghead{W.Roh et al.}{Overlap Region Aware Multi-view 3D Object Detection}

\def\etal{\emph{et al}\bmvaOneDot}
\newcommand{\myparagraph}[1]{\vspace{2pt}\noindent{\bf #1}}
\footnotetext{Corresponding Authors}

\usepackage{wrapfig}
\usepackage{tikz}
\usepackage{comment}
\usepackage{amsmath,amssymb}
\usepackage{color}
\usepackage{kotex}
\usepackage{booktabs}
\usepackage{adjustbox}
\usepackage[accsupp]{axessibility}

\usepackage{multirow}

\begin{document}

\maketitle
\vspace{-1.5em}
\begin{abstract}
Current multi-view 3D object detection methods often fail to detect objects in the overlap region properly, and the networks' understanding of the scene is often limited to that of a monocular detection network. Moreover, objects in the overlap region are often largely occluded or suffer from deformation due to camera distortion, causing a domain shift. To mitigate this issue, we propose using the following two main modules: (1) Stereo Disparity Estimation for Weak Depth Supervision and (2) Adversarial Overlap Region Discriminator. The former utilizes the traditional stereo disparity estimation method to obtain reliable disparity information from the overlap region. Given the disparity estimates as supervision, we propose regularizing the network to fully utilize the geometric potential of binocular images and improve the overall detection accuracy accordingly.
Further, the latter module minimizes the representational gap between non-overlap and overlapping regions. We demonstrate the effectiveness of the proposed method with the nuScenes large-scale multi-view 3D object detection data. Our experiments show that our proposed method outperforms current state-of-the-art models, i.e., DETR3D and BEVDet.

\end{abstract}


\vspace{-1em}
\section{Introduction}
\label{sec:intro}
\begin{figure*}[t]
    \centering
    \includegraphics[width=.9\linewidth]{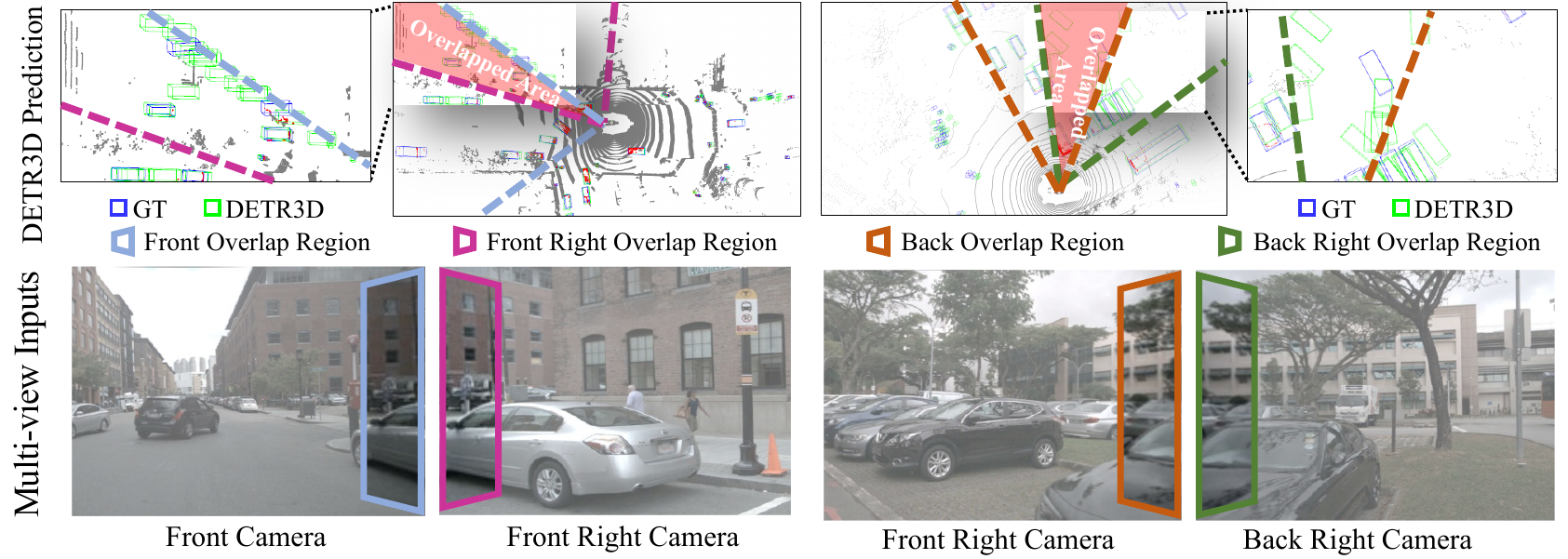}
    \caption{Examples where DETR3D~\cite{wang2022detr3d} fails to properly detect objects in the overlapped regions (see dotted line), resulting in performance degradation due to multiple false positives. The first row represents the bird's eye view of the two scenes in the second row (see there are more false positives in the overlapped regions than in others). The second row shows overlapped areas between multi-view images used as input to the network.
    }
    \label{fig:figure1}
    \vspace{-1.5em}
\end{figure*}

Object detection in 3D space plays a crucial role in various real-world applications, including autonomous driving systems. Existing 3D object detection methods~\cite{he2020svga,shi2020pv,deng2020voxel,shi2021pv,mao2021pyramid} based on point clouds from LiDAR sensors often yield reliable results, but these methods suffer from a large budget to establish LiDAR sensors per vehicle. Further, camera-based object detection methods~\cite{wang2021fcos3d,wang2022probabilistic,wang2022detr3d,li2020rtm3d} using monocular images are economical but, their performance is suboptimal due to insufficient depth cues. Stereo vision-based object detection methods~\cite{liu2021yolostereo3d, chen2020dsgn, wang2021plumenet, li2019stereo} might be an alternative option as they outperform monocular detection approaches with accurate depth estimation. Still, constraints in setting surround-view stereo vision systems need to be resolved. Recently, multi-view (and surround-view) camera systems have become an alternative balanced option as they can resolve some of the weaknesses of monocular and stereo vision systems for the 3D object detection task, potentially replacing LiDAR sensors. 

Existing camera-only 3D object detection methods~\cite{wang2019pseudo, you2020pseudo, wang2021fcos3d, wang2022probabilistic} have mainly focused on predicting accurate depth to improve performance. 
Although estimating precise depth significantly impacts accuracy, it remains challenging. Primarily, there is a problem that depth may not be represented adequately on the pixel (e.g., difficulty in dealing with distant objects on pixels and depth compounding error properly). A landmark work in the camera-only multi-view 3D object detection task is DETR3D~\cite{wang2022detr3d}. It introduces a promising multi-view detection pipeline that processes six images concurrently in an end-to-end manner, predicting all objects around simultaneously and implicitly utilizing rich information in the overlapping regions. Even though DETR3D performs reasonably well, we found that the network (without explicit guidance) does not totally use the geometric potentials of multi-view camera systems. Specifically, the network’s understanding of the scene could be limited to that of a monocular detection network, resulting in multiple false positives in the overlapped regions, as shown in Figure~\ref{fig:figure1}. Thus, how to deal with this issue to boost detection accuracy remains a crucial problem.

As reported by Chen~\etal~\cite{chen2020dsgn}, disparity supervision, which fully pilots the network by exploiting the strong association of binocular images, substantially improves detection performance. Inspired by this observation, we propose to use stereo disparity estimation techniques on the overlap region, which is between all adjacent camera pairs in the surround-view setting. Although this region is relatively small, it serves as a geometric link between two images. Consequently, we apply outputs from the traditional stereo disparity estimation model as weak depth supervision to improve the detection accuracy over the overlap region. We empirically found that this supervision significantly improves the overall detection accuracy.

This is only part of a story. Unlike the human vision system that quickly identifies an object across overlapped cameras, we empirically observe that DNN has a strong inductive bias toward identifying objects individually in each single-view image. This often results in failing to utilize additional information of the same instance appearing in other different-view images. Moreover, we observe a domain shift effect between the overlap region (i.e., the region far from the camera center) and the non-overlap region (i.e., the region near the camera center) due to camera lens distortion. Thus, we further propose to train an adversarial overlap region discriminator, which minimizes the domain gap between objects in the non-overlap regions vs. overlap regions. We validate from experiments that such adversarial training makes the overlap region performance more robust.

\begin{figure*}[t]
    \centering
    \includegraphics[width=.9\linewidth]{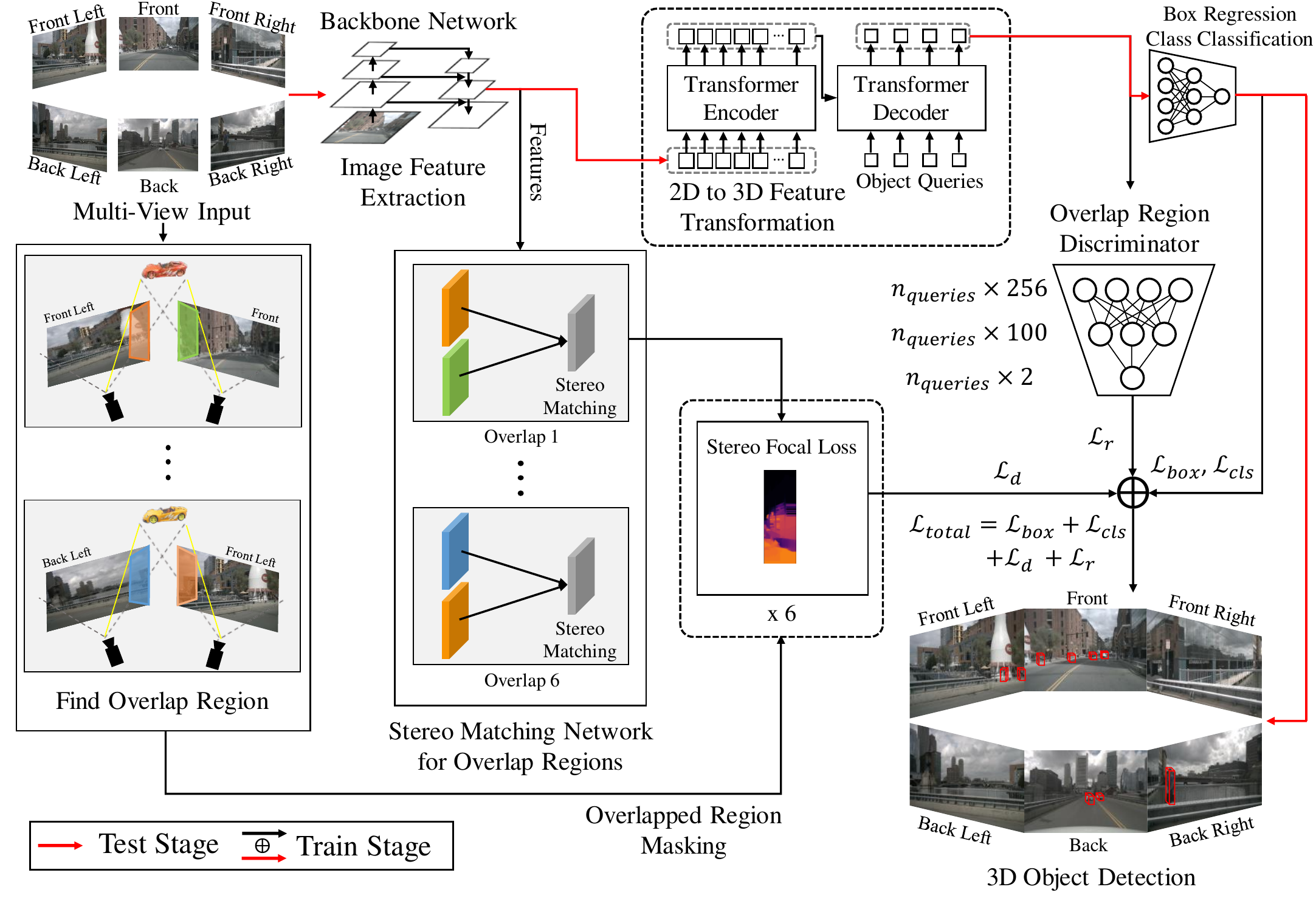}
    \caption{An overview of our proposed architecture. Built upon DETR3D~\cite{wang2022detr3d}, our model takes multi-view camera inputs and outputs a set of 3D bounding boxes for objects in the scene. Our model consists of two main modules: (1) Stereo Matching Network for Weak Depth Supervision, where our depth estimation head is trained to predict a dense depth map of the overlap region. The ground-truth depth map is obtained by a traditional stereo disparity estimation algorithm. (2) Adversarial Overlap Region Discriminator, which minimizes the gap between non-overlap regions vs. overlap regions, improving the overall detection performance.}
    \label{fig:figure2}
    \vspace{-1.5em}
\end{figure*}

We start from the state-of-the-art multi-view 3D object detection model called DETR3D~\cite{wang2022detr3d} as shown in Figure~\ref{fig:figure2}. Built upon DETR3D, we introduce the following two main modules: (i) Stereo Disparity Estimation for Weak Supervision and (ii) Adversarial Overlap Region Discriminator. We evaluate the effectiveness of our proposed method using the nuScenes~\cite{caesar2020nuscenes} dataset, which is a widely used large-scale multi-view 3D object detection benchmark. Through comprehensive experiments, we verify that our proposed model generally outperforms the state-of-the-art approaches in the camera-only 3D object detection task. Our main contributions are summarized as follows:
\vspace{-0.5em}
\begin{itemize}
    \item[•] We report that existing works often neglect properly dealing with objects in the overlap region, which limits fully using the geometric potentials of multi-view camera systems, causing performance degradation.
    \vspace{-.5em}
    \item[•] We propose to use outputs from the traditional stereo disparity estimation model on the overlap region and apply them as weak supervision to improve the detection accuracy over the overlap region. We empirically find that this supervision significantly improves the overall detection accuracy.
    \vspace{-.5em}
    \item[•] We introduce an overlap region discriminator that adversarially learns to minimize the covariate shift between objects from non-overlap regions vs. from overlap regions.
\end{itemize}
\vspace{-0.5em}
%

\vspace{-1em}
\section{Related Work}
\label{sec:related}

\myparagraph{3D Object Detection.}
As the study of inferring objects in 3D plane begins, Mono3D~\cite{chen2016monocular} conducts 3D object detection task using multiple information such as RGB images, object instruction segmentation, context, and location prior information. The model proposed by Roddick, T.~\etal~\cite{roddick2018orthographic} utilizes Bird-Eye-View(BEV) to compensate for performance degradation due to incorrect depth feature extracted from solely RGB images. MonoPSR~\cite{ku2019monocular} presents a method to significantly reduce 3D search space using 2D object detection and exploit point clouds to recover local shape and scale. FCOS3D~\cite{wang2021fcos3d} and RTM3D~\cite{li2020rtm3d} predicts 3D bounding box more accurately through features guided by geometric information. In addition, PGD~\cite{wang2022probabilistic} shows that geometric interactions between objects enhance the reliability of depth. Yet, the methods mentioned above perform detection tasks independently for each image. Thus, to expand for multiple cameras, each frame processing is performed before integrating the outputs in the post-processing stage. To address this problem,  DETR3D~\cite{wang2022detr3d} introduces a 3D object detection method that simultaneously processes multi-view images. DETR3D predicts 3D bounding boxes via backward geometric projection and operates set-to-set prediction without post-processing. Also, ImVoxelNet~\cite{rukhovich2022imvoxelnet} implements a surrounding view 3D object detector similar to the LiDAR system by optimizing the chronic computation of multiple images. Although they use multi-view images, they do not take advantage of the features of multiple camera settings. Hence, we focus on overlaying clues between images, which are multi-view characteristics, to improve detection performance.

In addition to the approaches mentioned above, recently, another solution of transforming the image features into BEV representation and applying it to 3D object detection has been widely studied. Lift-Splat-Shoot (LSS)~\cite{philion2020lift} introduces a view transform method that infers depth distribution and projects multiple image features into BEV representation. BEVDet~\cite{huang2021bevdet} and BEVDet4D~\cite{huang2022bevdet4d}, which extend LSS, demonstrate that applying the BEV features to 3D object detection is practical. Following BEVDet, BEVDepth~\cite{li2022bevdepth} constructs better BEV features with trustworthy depth prediction. BEVFormer~\cite{li2022bevformer} employs spatiotemporal cues by interacting with spatial and temporal space via predefined grid-shaped BEV queries. Note that our proposed method could potentially be applied to BEV-based approaches in an ad-hoc manner, as we focus on regularizing networks to deal with objects in the overlap region properly. However, we leave it as future work and will first focus on improving DETR3D-based approaches.

\myparagraph{Stereo 3D Object Detection.}
The stereo 3D object detection task that utilizes binocular information is similar to the human system. Inspired by depth estimation models~\cite{zbontar2016stereo,kendall2017end,chang2018pyramid}, DSGN~\cite{chen2020dsgn} proposes an end-to-end model that simultaneously uses plane-sweep volume and 3D geometric volume to predict 3D bounding boxes. PLUMEnet~\cite{wang2021plumenet} directly constructs a pseudo-LiDAR feature volume (PLUME) in 3D space. Reliable depth features extracted from the stereo view yield outstanding performance, whereas the 3D cost volume formed in the stereo network directs a lot of computation complexity. YOLOStereo3D~\cite{liu2021yolostereo3d}, a single-stage 3D detection network, effectively deals with the trade-off between computational complexity and depth accuracy with light-weight cost volume. Additionally, MobileStereoNet~\cite{shamsafar2022mobilestereonet} introduces a way to leverage MobileNets~\cite{howard2017mobilenets} to reduce the computation cost of deep networks without sacrificing accuracy. From these observations, our method utilizes the disparity information as 3D geometric cues in multi-view settings to enhance the accuracy of 3D object detection.



\vspace{-1em}
\section{ORA3D}
\label{sec:method}

In the following sections, we present a novel multi-view 3D object detection model that leverages rich information from a camera-only multi-view vision system. Our model is built upon the state-of-the-art DETR3D~\cite{wang2022detr3d} model, and we propose to use the following two main modules: Stereo Disparity Estimation for Weak Supervision (Section~\ref{sec:stereo}) and Adversarial Overlap Region Discriminator (Section~\ref{sec:adversarial}).

\vspace{-0.5em}
\subsection{Stereo Disparity Estimation for Weak Depth Supervision}\label{sec:stereo}
Existing work suggests that 3D object detectors from stereo vision can take advantage of estimating accurate depth for objects from binocular images. Such depth information is helpful for camera-only object detectors, which often lack reliable depth information. Our surround-view camera setting differs from the conventional stereo vision task -- only a tiny portion of the overlap region (i.e., less than 20\%) is available. 
In this work, we advocate for leveraging such overlap regions to supervise networks learning depth cues, potentially providing better 3D detection performance.

\begin{wrapfigure}{r}{0.56\textwidth}
    \vspace{-1em}
    \centering
    \includegraphics[width=\linewidth]{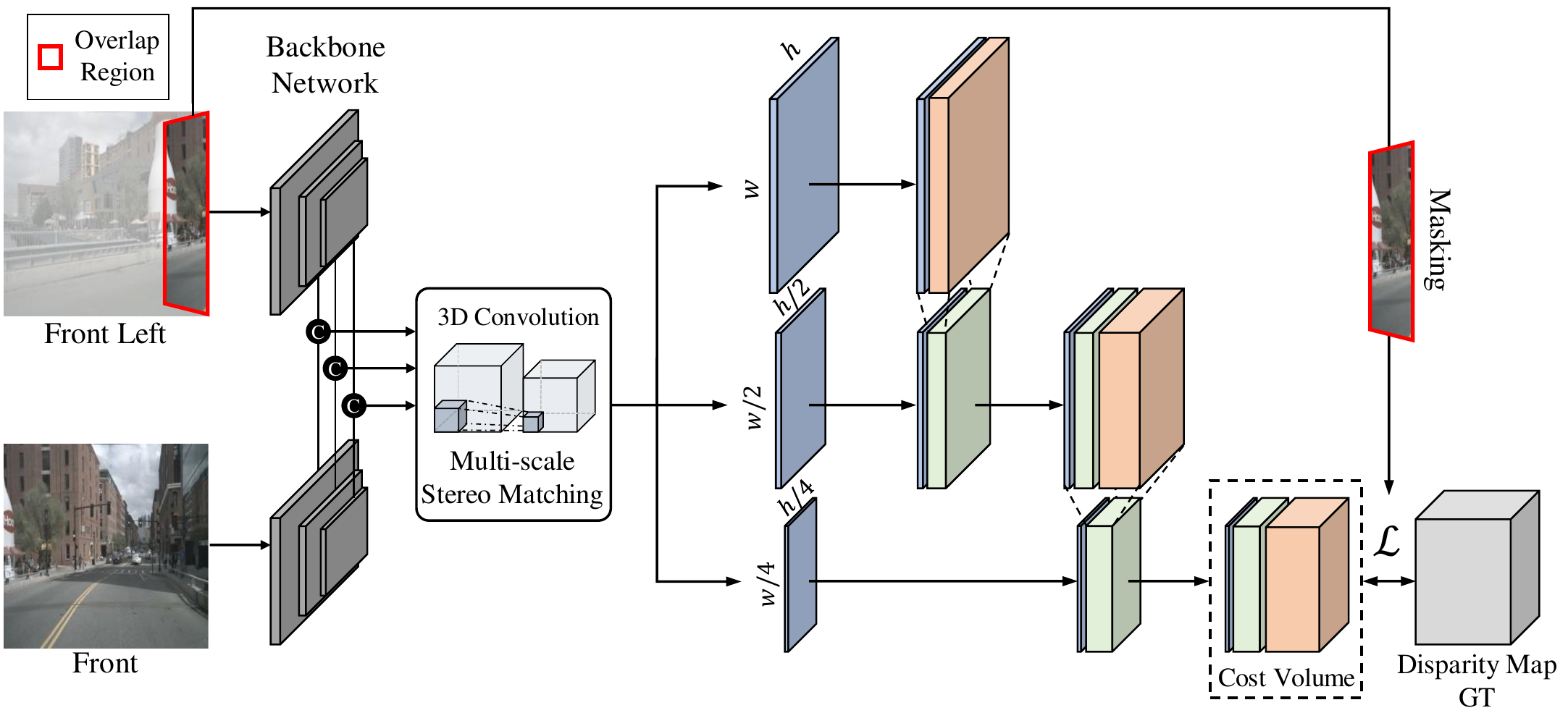}
    \caption{Following Liu~\etal~\cite{liu2021yolostereo3d}, our Stereo Disparity Estimation head is co-trained to compute the disparity map from two overlapped images. 
    }
    \label{fig:figure4}
    \vspace{-0.5em}
\end{wrapfigure}


\myparagraph{Learning Depth Cue by Multi-view Stereo Matching.}
When considering the multi-view camera system, adjacent cameras have a strong association. We regard this association comes from overlap regions and can be extended to geometric guides. To interactively supervise the network, we train the Stereo Disparity Estimation head, which reconstructs a dense disparity map with overlap region pairs of neighboring cameras. 

We follow the recent work by Liu~\etal~\cite{liu2021yolostereo3d} to implement the disparity estimation head. As illustrated in Figure~\ref{fig:figure4}, our stereo network extracts features of an image pair with a standard visual encoder. Our disparity estimation head outputs a cost volume through multi-scale stereo matching. To obtain the target disparity map, we use the output from the conventional stereo matching algorithm~\cite{hirschmuller2007stereo}, which performs pixel-wise mutual information-based matching. 

Further, we use a binary mask to consider losses from overlap regions, blocking gradients from non-overlap regions for training our Stereo Disparity Estimation module. Given the known internal camera parameters and external parameters, we first map the point $(x_s, y_s)$ in the source image coordinate to the point $(X, Y, Z)$ in the world coordinate, i.e., $(x_s, y_s)^\top\mapsto(X, Y, Z)^\top$. Then, we project the point $(X, Y, Z)$ back to the point $(x_t, y_t)$ in the target (or neighboring) camera coordinate frame, i.e., $(X, Y, Z)^\top\mapsto(x_t, y_t)^\top$. We provide more details in the supplemental material.


\myparagraph{Stereo Disparity Estimation Loss.} 
We use the following stereo focal loss~\cite{zhang2020adaptive} ${\mathcal{L}}_{d}$ to optimize our Stereo Disparity Estimation head:
\begin{equation}
    {\mathcal{L}}_{d} = \frac{1}{|\mathcal{I}_o|} \sum_{x\in\mathcal{I}_o}\sum_{d=0}^{D-1}((1-P_{x}(d)))^{-\alpha}(-P_{x}(d)\log{{\hat P_{x}(d)}})
\end{equation}
for pixels $x$ in the overlap region $\mathcal{I}_o$. Note that $d\in\{0, 1, \dots, D-1\}$ represents a discretized disparity, and $\alpha\in\mathcal{R}$ is the focus weight that is set to 1.0. $P_x(d)$ represents the target probabilistic distribution given a disparity $d$ for a pixel $x$ defined as follows: $P_x(d)=\textrm{Softmax}(-2|d-d^{gt}|)$ where $d^{gt}$ is the target (or ground-truth) disparity map. Similarly, $\hat P_{p}(d)$ is the predicted probabilistic distribution of a given disparity $d$.

%
%

\subsection{Adversarial Overlap Region Discriminator}\label{sec:adversarial}
In general, the human visual system can easily recognize objects at the edge of the image as well as in the center. Furthermore, it is not difficult to identify the same instances that appear simultaneously in different-view images. However, empirically, we discovered that DNN-based multi-view object detection models have a strong inductive bias toward identifying objects individually in each image. Note that objects in the overlap region are often occluded mainly due to limited Field Of View (FOV) of the camera or suffer from deformation due to camera lens distortion (e.g., pincushion distortion). Finally, we observe a domain shift effect between the overlap region (i.e., region far from camera center) vs. non-overlap region (i.e., region near camera center).

In addition to leveraging stereo disparity information, which gives an implicit bias to process the same object in different-view images together, we propose to use a regularizer to explicitly minimize the representational gap between non-overlap regions vs. overlap regions. Specifically, we constrain the object detection head from learning region-invariant information via an adversarial framework using Gradient Reversal Layer (GRL)~\cite{ganin2016domain}.

Given a query feature ${\bf q}_i$ for $i=\{1, 2, \dots, |Q|\}$ in the transformer (i.e., object detection decoder), an overlap region discriminator $f_d$ takes such query features. Formally, this discriminator needs to correctly predict its source region. Thus, this classifier $f_d$ is trained using region classification loss $\mathcal{L}_r$ as follows: ${\mathcal{L}}_{r} = -\mathbb{E}_{q, y_r\sim\mathbb{D}}\big[\sum_{r\in\mathcal{R}}y_{r} \log f_d(q)_{r}\big] $
where $\mathbb{E}_{q, y_r\sim\mathbb{D}}$ indicates an expectation over samples $(q, y_r)$, which are drawn from the (input) data distribution $\mathbb{D}$. Plus, $f_d$ is trained to classify whether a query feature is from the overlap region ($y_r$ is set to 1) or from the non-overlap region ($y_r$ is set to 0), so the output dimension from this module is 2 (i.e. $y_r\in\{0,1\}$). 
Notably, to reduce inductive bias, we design this ordinary discriminator as a special. We intentionally make a loss negative (i.e., $-\mathcal{L}_r$), preventing the discriminator from rightly distinguishing. Finally, our network is robust to all areas, minimizing bias that boosts performance degradation for overlapped regions.

\myparagraph{Loss Function.}
Ultimately, our model is trained end-to-end by minimizing the following loss function $\mathcal{L}_{\text{total}}$:
\vspace{-0.3em}
\begin{equation}
    {\mathcal{L}}_\text{total} = ~\lambda_{\text{cls}}{\mathcal{L}}_{\text{cls}}~+~\lambda_{\text{box}}{\mathcal{L}}_{\text{box}}~+~\lambda_{d}{\mathcal{L}}_{d}~-~\lambda_{r}{\mathcal{L}}_{r}
    \vspace{-0.3em}
\end{equation}
where $\lambda_{\text{cls}}$, $\lambda_{\text{box}}$, $\lambda_d$, and $\lambda_r$ are hyperparameters that are drawn from a grid search to control the strength of $\lambda_{\text{cls}}$, $\mathcal{L}_{\text{box}}$, $\mathcal{L}_{d}$, and $\mathcal{L}_{r}$, respectively.


\vspace{-1em}
\section{Experiments}
\label{sec:experiments}

\myparagraph{Dataset.} 
We use the nuScenes~\cite{caesar2020nuscenes} dataset, a large-scale multi-view object detection benchmark. The nuScenes dataset provides a full 360-degree field of view captured by six different viewing cameras. This comprises 20-second-long 1,000 video sequences, which are fully annotated with 3D bounding boxes for 10 object classes. The dataset covers 28k annotated samples for training, and validation and test contain 6k scenes each. 

\myparagraph{Evaluation Metrics.}
We follow the official evaluation protocol of nuScenes~\cite{caesar2020nuscenes}. We use a set of True Positive metrics (TP metrics) for each prediction that was matched with a ground-truth box. We employ the following 5 TP metrics: Average Translation Error(ATE), Average Scale Error(ASE), Average Orientation Error(AOE), Average Velocity Error(AVE), and Average Attribute Error(AAE). All TP metrics are also calculated using a 2m center distance threshold during matching, and they are all designed to be positive scalars. We also measure mean average precision (mAP). Lastly, we use the nuScenes Detection Score (NDS) to measures a consolidated scalar metric defined as follows: $\text{NDS} = {1 \over 10}[5~\text{mAP} + \sum_{\text{mTP} \in \mathbb{TP}}{(1 - \textnormal{min}(1, \text{mTP}))}]$ where $\mathbb{TP}$ is five TP metrics. Note that implementation and training details are provided in the supplemental material.

\begin{table*}[t]
\caption{
    Comparing our 3D object detector and the state-of-the-art on the nuScenes~\cite{caesar2020nuscenes} dataset.
    All methods based on camera modality. $\dagger:$ trained with CBGS~\cite{zhu2019class}. $\ast:$ initialized from pre-trained model on extra data. $\ddagger:$ initialized from DD3D checkpoint.
    }
    \begin{minipage}[t]{0.5\textwidth}
        \label{table:quantitative-nuscenes}
        \vspace{-0.5em}
        \resizebox{\textwidth}{!}{
        \begin{tabular}{lcccc}\toprule
                \parbox{2cm}{\centering Model} & ~Mono$|$Multi~ & ~Backbone & ~NDS($\uparrow$) & ~mAP($\uparrow$)  \\\midrule
                CenterNet~\cite{zhou2019objects} & Mono & DLA34 & 0.328 & 0.306 \\
                
                FCOS3D~\cite{wang2021fcos3d} & Mono & ResNet101 & 0.415 & 0.343  \\
                PGD~\cite{wang2022probabilistic} & Mono & ResNet101 & 0.428 & 0.369 \\\midrule
                DETR3D~\cite{wang2022detr3d} & Multi & ResNet101 & 0.425 & 0.346  \\
                DETR3D$^\dagger$~\cite{wang2022detr3d} & Multi & ResNet101 & 0.434 & 0.349  \\\midrule
                Ours & Multi & ResNet101 & \textbf{0.445} & \textbf{0.367}  \\
                \bottomrule
                \vspace{0.05em}
        \end{tabular}}
        \centering
        \scriptsize{(a) Validation set}
    \end{minipage}
    \hfill
    \begin{minipage}[t]{0.5\textwidth}
        \label{table:quantitative-nuscenes-test}
        \vspace{-0.5em}
        \resizebox{\textwidth}{!}{
        \begin{tabular}{lcccc}\toprule
                \parbox{2cm}{\centering Model}~ & ~Mono$|$Multi~ & ~Backbone & ~NDS($\uparrow$) & ~mAP($\uparrow$)  \\\midrule
                FCOS3D~\cite{wang2021fcos3d}~ & Mono & ResNet101 & 0.428 & 0.358 \\
                PGD~\cite{wang2022probabilistic} & Mono & ResNet101 & 0.448 & 0.386  \\
                DD3D$^\ast$~\cite{park2021dd3d} & Mono & V2-99 & 0.477 & 0.418  \\\midrule
                DETR3D$^\ddagger$~\cite{wang2022detr3d} & Multi & V2-99 & 0.479 & 0.412  \\
                BEVDet$^\ddagger$~\cite{huang2021bevdet} & Multi & V2-99 & 0.482 & 0.422  \\\midrule
                Ours$^\ddagger$ & Multi & V2-99 & \textbf{0.489} & \textbf{0.423} \\
                \bottomrule
                \vspace{0.05em}
        \end{tabular}}
        \centering
        \scriptsize{(b) Test set}
    \end{minipage}
\end{table*}

\myparagraph{Performance Comparison with SOTA.}
We compare our proposed ORA3D with existing state-of-the-art methods including CenterNet~\cite{zhou2019objects}, FCOS3D~\cite{wang2021fcos3d}, PGD~\cite{wang2022probabilistic}, DD3D~\cite{park2021dd3d}, DETR3D~\cite{wang2022detr3d}, and BEVDet~\cite{huang2021bevdet}. The first four approaches take multiple single-view images independently and combine detection outputs for the final output using non-maximum suppression (NMS). The last two approaches (DETR3D and BEVDet) are state-of-the-art multi-view  3D object detection models. As shown in Table~\ref{table:quantitative-nuscenes} (a) and (b), ORA3D generally outperforms other methods (compare the last row vs. others) in terms of (a consolidated metric) NDS and mAP both in validation and test data. Note that FCOS3D uses test-time augmentation and a customized data augmentation strategy, thus requiring more epochs and model ensembles. Though we do not use such an augmentation strategy, ours performs better than these methods. Note that we initialize with DD3D~\cite{park2021dd3d} pre-trained model and use the same backbone (V2-99~\cite{lee2020centermask}) for a fair comparison with DETR3D and BEVDet. 


\begin{table}[t]
    \begin{center}
    \caption{Average precision (AP) for each object class on the nuScenes~\cite{caesar2020nuscenes} test set. Higher value is better. \textit{Abbr.} C.V: construction vehicle, T.C: traffic cone.}
    \vspace{0.5em}
    \label{table:quan_ap}
    \resizebox{.95\textwidth}{!}{
    \begin{tabular}{lccccccccccc}\toprule
        \parbox{2cm}{\centering Model~} & ~Car~ & ~Truck~ & ~Bus~ & ~Trailer~ & ~C.V~ & ~Ped.~ & ~Motor.~ & ~Bicycle~ & ~T.C~ & ~Barrier~ & ~mAP\\\midrule
        DETR3D~\cite{wang2022detr3d} & 0.603 & 0.333 & 0.290 & \textbf{0.358} & 0.170 & 0.455 & 0.413 & 0.308 & 0.627 & 0.565 & 0.412\\
        Ours & \textbf{0.609} & \textbf{0.338} & \textbf{0.323} & 0.347 & \textbf{0.174} & \textbf{0.467} & \textbf{0.420} & \textbf{0.311} & \textbf{0.649} & \textbf{0.589} & \textbf{0.423}\\\bottomrule
    \end{tabular}}
    \vspace{-2em}
\end{center}
\end{table}

Further, we observe in Table~\ref{table:quan_ap} that ours show higher mAP scores in all objects except Trailer. Note that ours is initialized from the same backbone of DETR3D~\cite{wang2022detr3d}. This confirms that our proposed regularization terms clearly improve the overall detection performance.




\begin{table*}[t]
    \begin{center}
    \caption{Detection performance comparison with the state-of-the-art approaches for \textbf{objects in the overlap region}. nuScenes~\cite{caesar2020nuscenes} validation set are used and all use the same backbone. 
    }
    \vspace{0.5em}
    \label{table:quan_inversion}
    \resizebox{.95\textwidth}{!}{
    \begin{tabular}{lcccccccc}\toprule
        \parbox{2cm}{\centering Model} & Mono$|$Multi & NDS~($\uparrow$) & mAP~($\uparrow$) & mATE~($\downarrow$) & mASE~($\downarrow$) & mAOE~($\downarrow$) & mAVE~($\downarrow$) & mAAE~($\downarrow$) \\\midrule
        FCOS3D~\cite{wang2021fcos3d} & Mono & 0.317 & 0.213 & 0.841 & \textbf{0.276} & 0.604 & 1.122 & \textbf{0.173} \\\midrule
        DETR3D~\cite{wang2022detr3d} & Multi & 0.356  & 0.231 & 0.825 & 0.280 & 0.400 & 0.863 & 0.223 \\
        Ours & Multi & \textbf{0.408}  & \textbf{0.264} & \textbf{0.677} & 0.280 & \textbf{0.361} & \textbf{0.746} & 0.181 \\\bottomrule
    \end{tabular}}
    \vspace{-1em}
\end{center}
\end{table*}

\myparagraph{Performance Comparison in Overlap Region.} 
We compare our proposed ORA3D again with existing state-of-the-art methods (FCOS3D~\cite{wang2021fcos3d}, and DETR3D~\cite{wang2022detr3d}), but we now focus on overlap regions. As shown in Table~\ref{table:quan_inversion}, our method outperforms FCOS3D and DETR3D in terms of all metrics except mASE and mAAE. This confirms that our overlap region aware approach effectively deals with objects in the overlap region, resulting in a large performance gain. For a fair comparison, all models use a ResNet101-based backbone, and only FCOS3D uses an augmentation strategy.

\begin{figure*}[t]
    \centering
    \includegraphics[width=\linewidth]{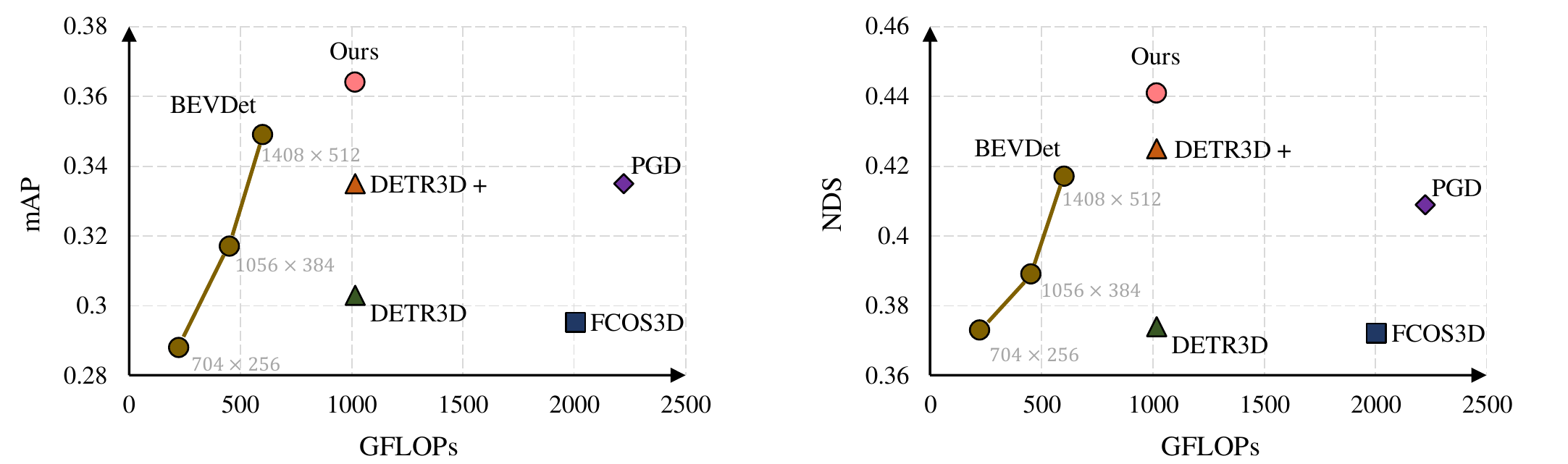}
    \caption{The computing budget and performance of different paradigms on the nuScenes~\cite{caesar2020nuscenes} validation set. Note that DETR3D$+$ indicates a DETR3D~\cite{wang2022detr3d} model with CBGS~\cite{zhu2019class}.} 
    \label{fig:gflop}
\end{figure*}

\begin{figure}[t]
    \centering
    \includegraphics[width=\linewidth]{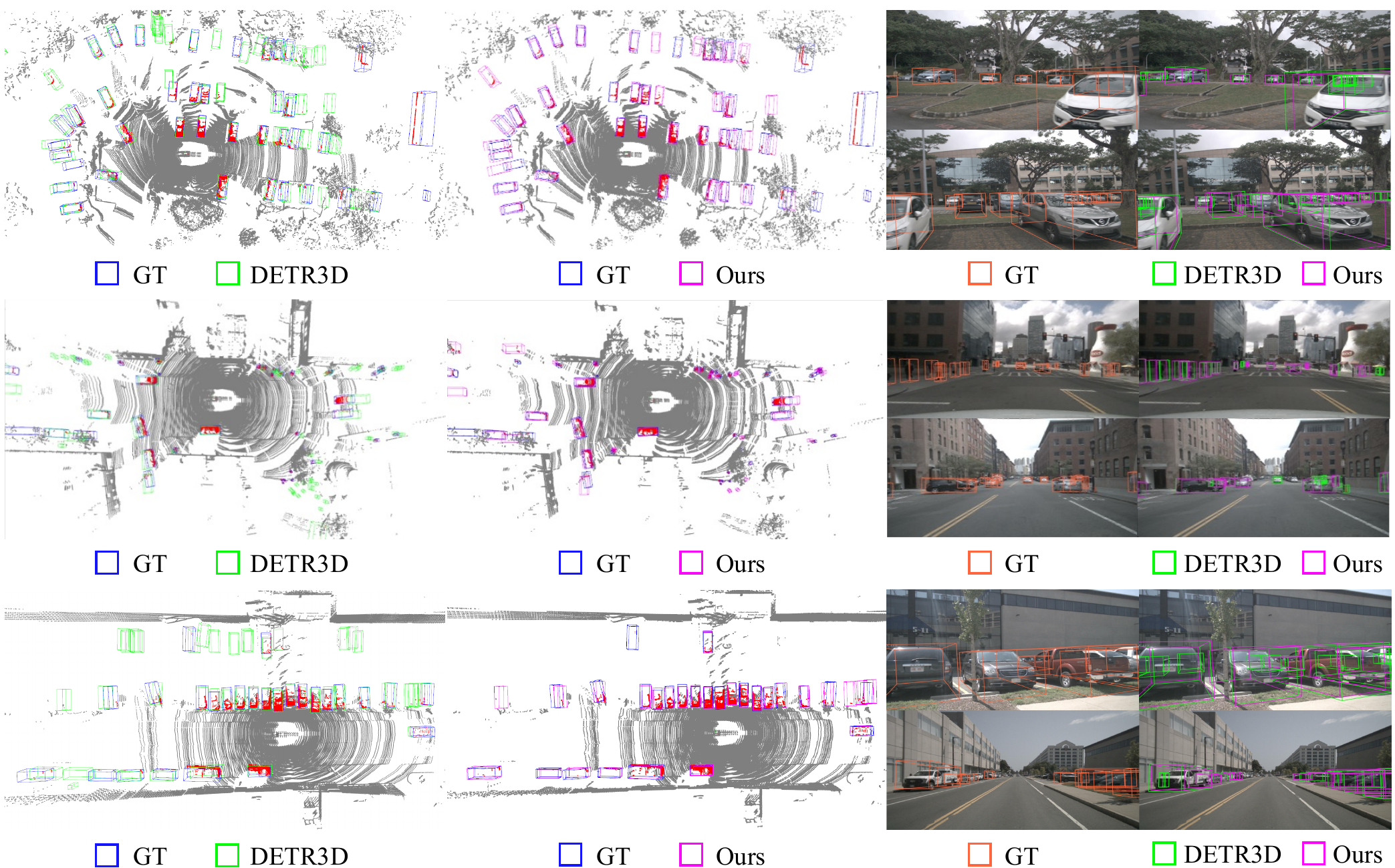} 
    \caption{Qualitative comparison between our method (purple) and DETR3D~\cite{wang2022detr3d} (green). Detected 3D bounding boxes for objects in the scene are projected into a bird's eye view perspective (left) and the image plane (right). See DETR3D produces more false positives in the overlapped regions than ours. }
    \label{fig:figure5}
    \vspace{-1.5em}
\end{figure}

\myparagraph{Qualitative Analysis.}
Existing CNN-based methods benefit from inductive bias even with relatively small datasets. However, large-scale datasets have become more common, and in these environments, inductive bias easily overfits specific datasets and causes various domain shift issues. In this section, we demonstrate that our proposed methods address these concerns while maximizing the capacity of multi-view camera systems.

Fig.~\ref{fig:figure5} shows the visualized results of 3D bounding boxes predicted by DETR3D (see green boxes) and our proposed method (see pink boxes). The ground-truth bounding boxes (see blue boxes) are overlaid. We project the predicted and ground-truth bounding boxes in the BEV perspective. In general, DETR3D and our proposed method generate reasonable results. However, the difference in the overlap region is apparent between the two methods, where DETR3D exhibits a relatively large number of false positive detections. Especially, our model is more robust for truncated or distorted objects across the entire region. Overall, our simple and effective methods amplify accuracy both qualitatively and quantitatively.

\myparagraph{Analysis of Computations.}
We compare our proposed method with existing 3D object detection methods, including FCOS3D~\cite{wang2021fcos3d}, PGD~\cite{wang2022probabilistic}, BEVDet~\cite{huang2021bevdet} and DETR3D~\cite{wang2022detr3d}. As illustrated in Fig.~\ref{fig:gflop}, our method outperforms other methods while requiring only a computational budget comparable to DETR3D. Furthermore, DETR3D uses CBGS~\cite{zhu2019class}, a helpful strategy for more balanced data distribution, to achieve higher performance. This strategy, however, alleviates the problem of data imbalance but requires more than four times the training time. Our method without CBGS demands less training time, but has the most impressive performance in Fig~\ref{fig:gflop}. Additionally, although another great work, BEVDet, uses fewer FLOPs by using smaller-sized images, ours shows higher scores (NDS: 0.445, mAP: 0.364). It would be worth exploring as a future work applying our approach to BEV-based models for potential performance improvement.

\newlength{\oldintextsep}
\setlength{\oldintextsep}{\intextsep}
\setlength\intextsep{0pt} 
\begin{wraptable}{r}{0.6\linewidth}
    \caption{Ablative analysis of our methods and DETR3D~\cite{wang2022detr3d} on the nuScenes~\cite{caesar2020nuscenes} {\it mini} validation set for the whole and overlap regions.}
    \vspace{0.5em}
    \label{table:ablation-full}
    \resizebox{\linewidth}{!}{
        \begin{tabular}{lcccc}\toprule
            \multirow{2}{*}{Model} & \multicolumn{2}{c}{Whole Region} & \multicolumn{2}{c}{Overlap Region}\\\cmidrule{2-5}
             & NDS~($\uparrow$) & mAP~($\uparrow$) & NDS~($\uparrow$) & mAP~($\uparrow$) \\
            \noalign{\smallskip}
            \hline
            \noalign{\smallskip}
            A. DETR3D~\cite{wang2022detr3d} & 0.338 & 0.247 & 0.266 & 0.151 \\\midrule
            B. A + $\mathcal{L}_d$ & 0.348 & 0.273 & 0.276 & 0.195 \\
            C. A + $\mathcal{L}_{r}$ & 0.339 & 0.286 & 0.274 & \textbf{0.215}\\
            D. A + $\mathcal{L}_d$ + $\mathcal{L}_{r}$ (Ours) & \textbf{0.353}  & \textbf{0.288} & \textbf{0.280} & 0.207\\\bottomrule
        \end{tabular}}
\end{wraptable}

\myparagraph{Ablation Study.}
We evaluate the variants of our method with DETR3D with and without the following two main loss terms: (i) $\mathcal{L}_d$: Stereo Disparity Estimation loss, (ii) $\mathcal{L}_r$: Adversarial Overlap Region Discrimination loss. In Table~\ref{table:ablation-full}, we quantitatively analyze the importance of our proposed methods for the whole region and the overlap region. Note that we use the nuScenes mini validation set. In Table~\ref{table:ablation-full} Whole Region, we observe that adding either $\mathcal{L}_d$ improves most of the metrics, while a similar trend is observed by adding $\mathcal{L}_r$ as well. Indeed, the stereo disparity loss is reflected during training to induce semantic geometric potentials evenly in all areas without distinguishing between overlap and non-overlap regions. In Table~\ref{table:ablation-full} Overlap Region, the effect of each component on the overlap region is further confirmed. Most importantly, we discover the remarkable efficiency of $\mathcal{L}_r$ in the target domain. The adversarial loss completely addresses the domain shift effect by deceiving the model into not correctly discriminating which region objects belong to. As a result, we confirm that our components improve overall performance.

\begin{wrapfigure}{r}{0.6\textwidth}
    \centering
    \caption{t-SNE visualization~\cite{van2008visualizing} of non-overlapped (blue) and overlapped (red) regions' features. We discover that our Overlap Region Discriminator suitably overcomes the domain shift effect between two regions. Best viewed in color.}
    \vspace{0.5em}
    \includegraphics[width=\linewidth]{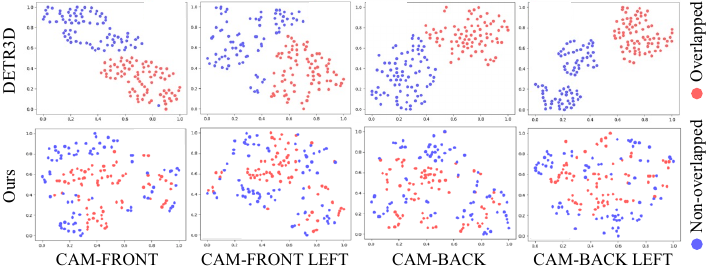}
    \label{fig:figure6}
    \vspace{-0.5em}
\end{wrapfigure}

\myparagraph{Analysis of Domain Shift Effect.}
We suppose that the multi-view 3D object detection method, such as DETR3D~\cite{wang2022detr3d}, without consideration of lens distortion and limited FOV, is likely to have an inductive bias between central and outer regions, causing a domain shift effect. To find the degree of inductive bias, we visualize the t-SNE~\cite{van2008visualizing} results of central (non-overlapped, blue) and edge (overlapped, red) regions' features in Fig~\ref{fig:figure6}. As we clearly see in the first row (DETR3D), the features of each two groups form distinguishable clusters. Our method starts with this observation. The overlap regions in multi-view images are clues to improving the overall performance. Ultimately, the features of our method's two groups (second row) are distributed harmoniously, minimizing bias. This distribution indicates that our proposed Overlap Region Discriminator successfully minimizes the representational gap.

\vspace{-1em}
\section{Conclusion}
\label{sec:conclusion}
In this paper, we present a novel pipeline to enhance the performance of 3D object detection using the 3D geometric cues in multi-view settings. We recognize that the overlaying region between images has become a weakness due to issues such as disconnected information and overlapping objects. However, the overlap region is also a novel material with the potential to improve overall performance. We develop approaches to reasonably use this small but highly informative area. The first is a light and robust stereo disparity estimation network for small overlap regions. This module allows the network to fully utilize the geometric potential of binocular images. Next is the adversarial overlap region discriminator, which is adversarially trained to minimize the gap between non-overlap regions and overlap regions. Overall, the experiments demonstrate that our two methods successfully work on detection accuracy. 


\section*{Acknowledgement}
\label{sec:acknowledgement}
This work is supported by Autonomous Driving Center, Hyundai Motor Company R\&D Division. S. Kim, W. Roh, G. Chang, and C. Kim are partially supported by the National Research Foundation of Korea grant (NRF-2022R1F1A1074334) and Institute of Information \& communications Technology Planning \& Evaluation (IITP) grant funded by the Korea government(MSIT) (No. 2019-0-00079, Artificial Intelligence Graduate School Program(Korea University). S. Moon and J. Kim are partially supported by the ICT Creative Consilience program (IITP-2022-2022-0-01819) and the ITRC(Information Technology Research Center) support program (IITP-2022-RS-2022-00156295).


\bibliography{egbib}
\end{document}